\documentclass[letterpaper, 10 pt, conference]{ieeeconf}  

\IEEEoverridecommandlockouts                              

\usepackage{graphics} 
\usepackage{epsfig} 
\usepackage{mathptmx} 
\usepackage{times} 
\usepackage{amsmath} 
\usepackage{amssymb}  
\usepackage{booktabs}
\usepackage{multirow}
\usepackage{capt-of}
\title{\LARGE \bf
SPICE: Stochastic Policy for Interaction in Cluttered Environments
}

\author{Nidhi Parayil, Benjamin Greenland, Thierry Peynot, and Chris Lehnert 
\thanks{The authors would like to thank the QUT Centre for Robotics and the QUT Research Engineering Facility (REF) team for their technical support and assistance throughout this research. Corresponding author: Nidhi Parayil (parayil@qut.edu.au).}
}

\begin{document}

\maketitle
\thispagestyle{empty}
\pagestyle{empty}

\begin{abstract}
Agricultural tasks such as harvesting and pruning require robots to reach through dense foliage, where contact is often unavoidable and vision is obstructed. Intermittent contact creates an imbalance between free-space and contact-rich observations, complicating the learning of contact-responsive behaviours. We present SPICE, a tactile-conditioned diffusion policy for reaching targets in dense plant canopies using tactile sensing as the sole environmental feedback. SPICE exhibits three behaviours: moving through compliant foliage, manoeuvring around partially obstructing stems, and manipulating branches that block the target path. To the best of our knowledge, SPICE is the first tactile-conditioned diffusion policy for target reaching through physically cluttered plant canopies. Across over 180 real-robot trials, SPICE was evaluated against tactile-conditioned, image-conditioned, and reactive-control baselines. Spatial tactile information and explicit contact conditioning proved important, with SPICE achieving \(95\%\) branch-manoeuvring success versus \(45\%\) for the strongest tactile baseline. SPICE also achieved branch manipulation beyond the reactive controller and remained effective under visual occlusion, where the image-conditioned DP failed. Zero-shot transfer across 40 trials on unseen tomato and lime plants achieved \(87.5\%\) success rate. These results demonstrate the potential of tactile-conditioned generative policies for interaction in visually occluded plant environments.
\end{abstract}

\section{INTRODUCTION}

\begin{figure}[t]
    \centering
    \includegraphics[width=0.48\textwidth]{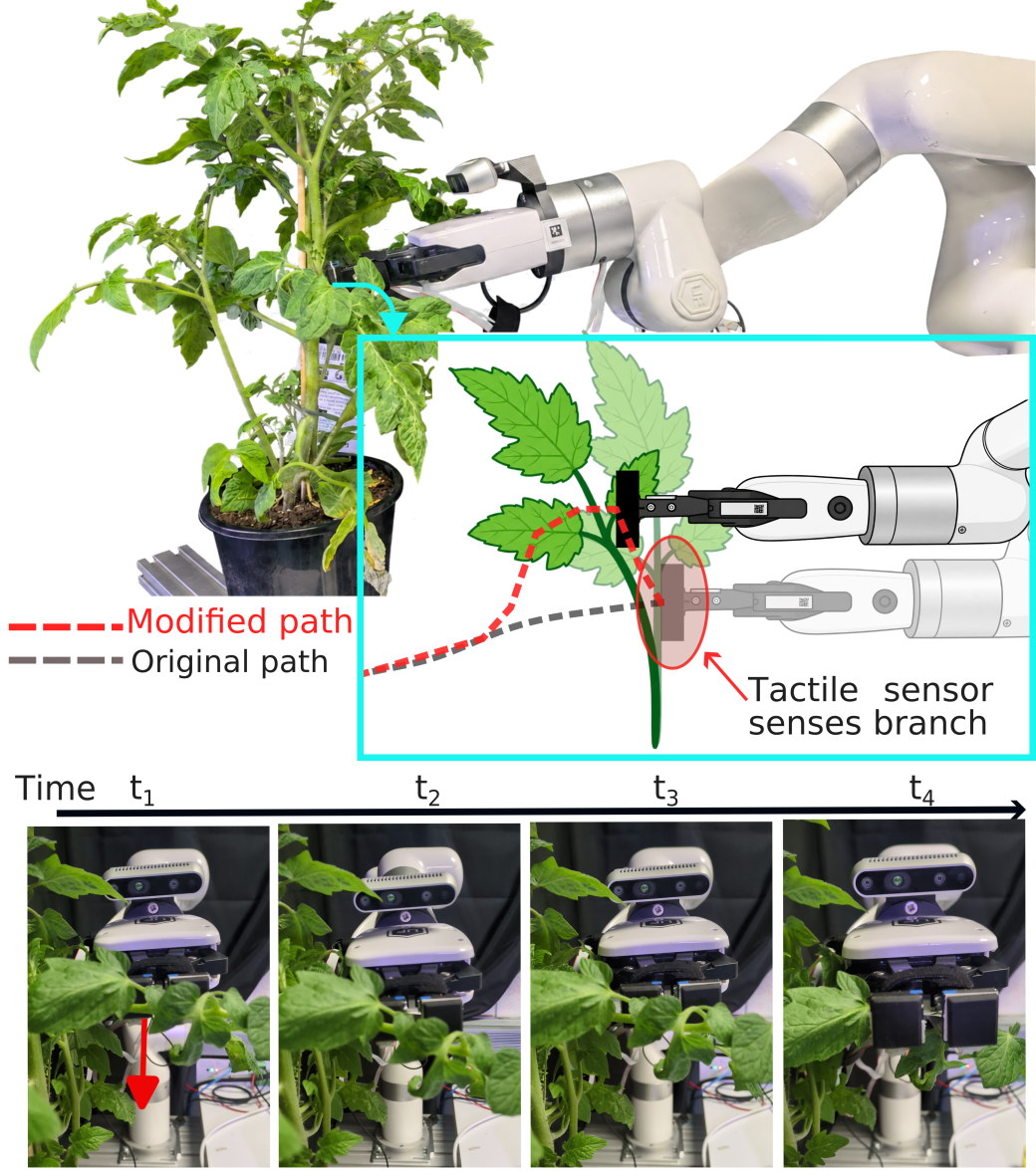}
    \vspace{-10pt}
\caption{Contact-responsive motion adaptation of SPICE in a cluttered tomato canopy. The blue inset shows tactile-driven trajectory adaptation, with the red and gray dashed lines indicating the modified and original paths, respectively. Tactile sensing enables SPICE to adapt its motion. The bottom row shows the interaction sequence from the tomato plant experiment: the end effector contacts the branch ($t_1$), pushes it downward ($t_2$), moves above the displaced branch ($t_3$), and continues toward the target ($t_4$).}
    \label{fig:tomato_fig}
    \vspace{-20pt}
\end{figure}

Agricultural labour shortages are increasing demand for robotic systems~\cite{barreiros2025careful}, yet many high-value crops, such as sweet pepper, berries, and citrus, remain difficult to automate because of dense, unstructured foliage~\cite{nazari2023deep}. Reaching a fruit or pruning site within dense foliage often requires moving through leaves and branches with unknown geometry, stiffness, and compliance. These structures create \textit{physical obstruction} by restricting access to the target and \textit{visual occlusion} by blocking the robot's view. When multiple nearby structures constrain the available path, the environment becomes \textit{cluttered}, making contact difficult to avoid and requiring the robot to adapt its motion using local physical feedback.\\
Humans handle such situations naturally by moving through compliant foliage, manoeuvring around partially obstructing stems, and displacing branches when there is insufficient space to move around them. In contrast, robotic interaction often relies on predefined contact responses or interaction models, limiting behavioural flexibility. Learning from demonstrations (LfD) offers a way to acquire diverse behaviours without explicitly modelling each interaction. However, tactile-based LfD has largely focused on gripper-level manipulation tasks such as grasping~\cite{ablett2024multimodal,zhou2023branch}, rather than guiding whole-arm motion through clutter. Extending tactile LfD to cluttered reaching is challenging because contact is intermittent and the required response changes as the robot progresses through the environment.\\
Existing methods for moving through plant canopies rely mainly on collision avoidance, model-based interaction, or predefined reactive contact responses~\cite{fu2026push,you2022precision}. Learning multiple contact-responsive behaviours directly from tactile feedback remains underexplored. We therefore introduce \textbf{SPICE - Stochastic Policy for Interaction in Cluttered Environments}, a vision-free, tactile-conditioned diffusion policy for reaching targets in occluded plant canopies. SPICE generates action sequences from tactile feedback and replans as contact evolves, enabling multiple contact-responsive behaviours within a single learned policy. Figure~\ref{fig:tomato_fig} illustrates SPICE displacing an obstructing branch in a tomato plant.

The main contributions of this work are:
\begin{itemize}
    \item To the best of the authors' knowledge, the first \textbf{tactile-conditioned diffusion policy enabling multiple contact-responsive} behaviours for reaching a target in visually occluded plant canopies.
    \item \textbf{A tactile-conditioning design for intermittent contact}, combining spatial taxel encoding with explicit contact information to generate whole-arm joint-space responses as interaction evolves.
    \item \textbf{Extensive real-robot validation and zero-shot transfer}, comprising over 180 controlled trials against  tactile-conditioned DP, image-conditioned DP, and reactive-control baselines, followed by 40 zero-shot trials on unseen lime and tomato plants.
\end{itemize}

\section{Related works}
Reaching through clutter requires moving towards a target while adapting to contact. Hybrid controllers combine position and force control using predefined objectives, while Model Predictive Controllers (MPC) optimise motion using estimated robot--environment dynamics~\cite{killpack2013fast,killpack2016model}. Such dynamics are difficult to model accurately in deformable environments such as plant canopies. Reactive controllers avoid explicit dynamics by responding directly to contact, either through predefined behaviour switching~\cite{brouwer2024tactile} or by combining target and contact objectives, as in RICE~\cite{rice}. However, incorporating multiple interaction responses can increase controller complexity and introduce conflicting objectives or local minima.\\
In agricultural robotics, predefined motions have been used to displace obstacles~\cite{xiong2020push}, reveal targets~\cite{gursoy2024occlusion,yao2025safe}, and regulate pruning contact~\cite{you2022precision,jain2025learning}. Model-based approaches plan branch interactions using spring models~\cite{jacob2024learning}, mechanical properties~\cite{rijal2025force}, or deformation parameters~\cite{rijal2026modelingbranchesactivemanipulation}. Reinforcement Learning (RL) has also enabled gentle manipulation~\cite{jacob2024gentle} and avoidance or displacement of plant structures~\cite{fu2026push}. These approaches still require manually specified motions, models, rewards, or interaction objectives. Learning from demonstration offers an alternative without explicit modelling. However, existing tactile agricultural methods mainly detect branch contact~\cite{zhou2023branch} or regulate predefined pushing motions~\cite{nazari2023deep}, rather than learning multiple contact-responsive behaviours for reaching through clutter.\\
Vision-language-action (VLA) models and Diffusion Policies (DPs) have shown strong performance in vision-based manipulation~\cite{chi2023diffusion,liu2024diff,ma2024hierarchical}, but have rarely been demonstrated under severe visual occlusion. Tactile-conditioned VLA and DP methods incorporate contact feedback~\cite{zhao2025polytouch,helmut2025tactile,ablett2024multimodal,yu2023mimictouch}, while TacDiffusion~\cite{wu2025tacdiffusion}, Reactive Diffusion Policy~\cite{xue2025reactive}, and Tube Diffusion Policy~\cite{xue2026tube} demonstrate tactile-conditioned generation for insertion, assembly, and local manipulation. However, these methods primarily address gripper-level interaction rather than whole-arm reaching through deformable clutter. Spatial tactile representations have also been explored using tactile skins such as DexSkin~\cite{wistreich2025dexskin}, but much of this work relies on camera-based tactile sensing. Tactile encoders for compact non-camera-based sensors, such as magnetic taxel arrays, remain less explored.\\
Taken together, existing methods either prescribe interaction responses or apply tactile learning primarily to local gripper manipulation. Learning multiple contact-responsive behaviours for whole-arm motion through clutter from spatial tactile feedback under visual occlusion remains underexplored. SPICE addresses this gap with a vision-free, tactile-conditioned diffusion policy that learns multiple interaction behaviours within a single policy.
\section{Methodology}
Moving through a cluttered plant canopy requires different contact-responsive behaviours depending on how surrounding structures obstruct the path. We first formulate this problem, then present SPICE, which combines spatial tactile encoding with diffusion-based joint-space action generation, followed by the demonstration collection and experimental setup.

\subsection{Problem Formulation}
We consider a manipulator reaching a specified target within a visually occluded and physically obstructed plant canopy. The canopy geometry, arrangement, stiffness, and contact dynamics are unknown. The robot observes its current configuration \(\mathbf q_t\), target configuration \(\mathbf q_{target}\), and tactile tensor \(\mathbf T_t\) from sensors mounted on the End Effector (EE); no visual observations or environmental model are provided.
\begin{figure}[!h]
    \centering
    \includegraphics[width=0.48\textwidth,trim={0 1.5cm 0cm 0cm},clip]{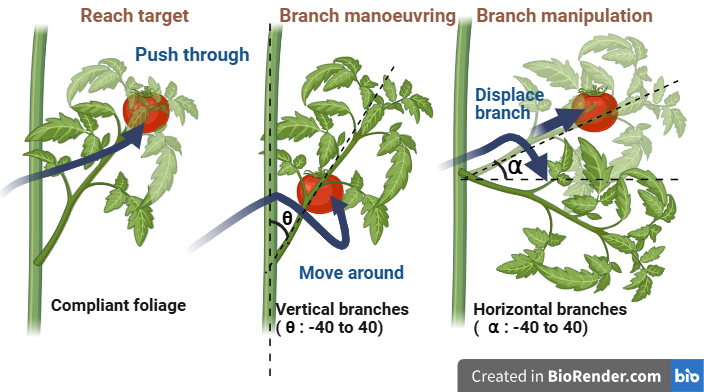}
    \vspace{-20pt}
    \caption{Contact-dependent behaviours for reaching an occluded target
(red): direct motion towards target through compliant foliage, manoeuvring around an
obstructing branch, and displacing a branch away from the target path.
Blue arrows indicate the end-effector trajectories.}
    \label{fig:interaction_behaviours}
    \vspace{-10pt}
\end{figure}
Prior work on reactive tactile control has demonstrated reliable target reaching and manoeuvring around partially obstructing stems~\cite{rice}. However, these local contact responses are insufficient when a branch fully blocks the target path and leaves no space to manoeuvre around it; in such cases, the robot must manipulate the environment to create access. We therefore consider three contact-dependent behaviours (Fig.~\ref{fig:interaction_behaviours}):
\begin{itemize}
    \item \textbf{Target reaching:} Moving towards the target through compliant foliage;
    \item \textbf{Branch manoeuvring:} Moving to the target while moving around a partially obstructing branch; and
    \item \textbf{Branch manipulation:} displacing a branch that fully obstructs the path before continuing towards the target.
\end{itemize}

To represent different environmental structures, we use branch configurations that create increasing levels of obstruction. With only compliant foliage and no obstructing branch, the robot can move directly towards the target. A vertical branch provides lateral free space, allowing the robot to manoeuvre around it without deliberately displacing the branch. In contrast, a horizontal branch blocks the available lateral path, making manoeuvring infeasible under the robot’s kinematic and sensing constraints. Because moving above the branch increases the likelihood of contact with unsensed robot surfaces, we use downward displacement as the manipulation behaviour. This creates access to the target while keeping the interaction near the tactile end effector.

We learn these behaviours with a single closed-loop policy from human teleoperation demonstrations, without providing behaviour labels. Given recent robot and tactile observations, the policy must move towards the target while adapting its contact response to different branch orientations and levels of obstruction.

\subsection{Spatial Tactile Encoder and Contact Conditioning}
\begin{figure}[h]
    \centering
    \includegraphics[width=.45\textwidth]{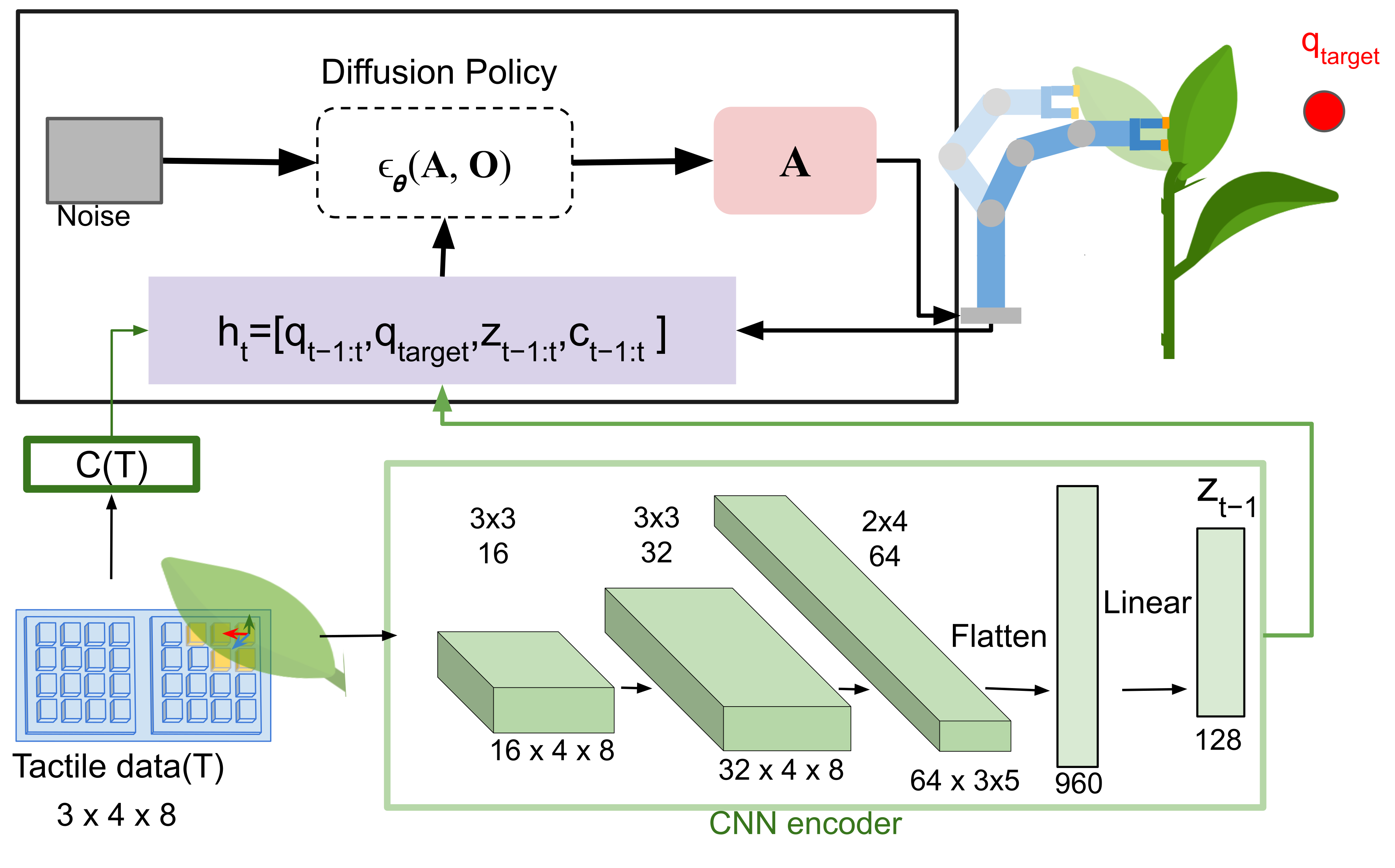}
        \vspace{-10pt}
\caption{ SPICE is a tactile-conditioned diffusion policy in which a CNN encodes the three-axis tactile tensor, which is combined with explicit contact conditioning and the current and target robot configurations to condition joint-space action generation.}
    \label{fig:method}
        \end{figure}
For reaching through clutter, the appropriate response depends not only on contact magnitude but also on how contact is distributed across the tactile surface. Preserving the spatial arrangement of the taxels therefore allows the policy to distinguish contact configurations that may require different motions. SPICE represents tactile feedback using a spatial embedding together with an explicit contact signal, as illustrated in Fig.~\ref{fig:method}.

Each tactile sensor contains a \(4\times4\) array of three-axis taxels. Measurements from the two sensors are concatenated into \(\mathbf{T}_t\in\mathbb{R}^{3\times4\times8}\), with channels corresponding to the \(x\)-, \(y\)-, and \(z\)-axis measurements. We use a CNN encoder with two \(3\times3\) convolutional layers of 16 and 32 channels, followed by a \(2\times4\) layer with 64 channels. All layers use unit stride and ReLU activation, with unit padding in the first two layers. The resulting \(64\times3\times5\) feature map is flattened and linearly projected to a 128-dimensional tactile embedding, \(\mathbf{z}^{\mathrm{tac}}_t\), trained end-to-end with the diffusion policy.

Because tactile contact occurs in only about \(30\%\) of the demonstrations, the policy must distinguish free-space motion without contact and physical interaction. We therefore complement the spatial embedding with an explicit contact signal \(c_t\), which indicates contact presence and intensity.\\
For each taxel measurement \(\mathbf{f}_{i,t}\), it is computed as
\[
c_t=\ln\!\left(
1+\max\left\{
\max_i\|\mathbf{f}_{i,t}\|_2-\tau,\,0
\right\}
\right),
\]
where \(\tau\) is the no-contact noise threshold. Thus, \(c_t=0\) in free space and increases with contact intensity.

\subsection{Conditional Diffusion Policy}
 
Unlike image-conditioned Diffusion Policy, SPICE conditions action generation on spatial tactile information, explicit contact state, and robot and target configurations, enabling operation under visual occlusion (Fig~\ref{fig:method}). The policy must infer the appropriate response from tactile interaction and generate coordinated joint motion as that interaction evolves. Using a two-step observation history, the conditioning vector is
\[
\mathbf{h}_t=
\operatorname{concat}\left(
\mathbf{q}_{t-1:t},
\mathbf{q}_{\mathrm{target}},
\mathbf{z}^{\mathrm{tac}}_{t-1:t},
c_{t-1:t}
\right),
\]
where \(\mathbf{h}_t\) comprises the robot state \(\mathbf{q}_t\in\mathbb{R}^n\), target state \(\mathbf{q}_{\mathrm{target}}\in\mathbb{R}^n\), tactile embedding \(\mathbf{z}^{\mathrm{tac}}_t\), and contact signal \(c_t\).

SPICE predicts a \(T_p=16\)-step sequence of absolute joint positions, $\mathbf{A}^{0}_t= [\mathbf{a}_t,\ldots,\mathbf{a}_{t+T_p-1}] \in\mathbb{R}^{T_p\times n}, \qquad \mathbf{a}_t\in\mathbb{R}^{n},$ where \(n=6\). The policy models $\mathbf{A}^{0}_t\sim p_{\theta}\!\left(\mathbf{A}^{0}_t\mid\mathbf{h}_t\right)$. Joint-space actions match the teleoperation demonstrations and robot-control interface. We adopt the conditional denoising formulation and temporal U-Net of Diffusion Policy~\cite{chi2023diffusion}, with \(\mathbf{h}_t\) incorporated through feature-wise linear modulation (FiLM).

During training, demonstrated action sequences are corrupted with Gaussian noise, $\mathbf{A}^{k}_t
=
\sqrt{\bar{\alpha}_k}\mathbf{A}^{0}_t
+
\sqrt{1-\bar{\alpha}_k}\boldsymbol{\epsilon},
\qquad
\boldsymbol{\epsilon}\sim\mathcal{N}(\mathbf{0},\mathbf{I}),$ where \(k\) denotes the diffusion step and \(\bar{\alpha}_k\) is defined by the noise schedule. The temporal U-Net \(\boldsymbol{\epsilon}_{\theta}\) predicts the added noise conditioned on \(\mathbf{A}^{k}_t\), \(k\), and \(\mathbf{h}_t\). The tactile encoder and denoising network are trained jointly using
\[
\mathcal{L}_{\mathrm{DP}}
=
\mathbb{E}_{\mathbf{A}^{0}_t,k,\boldsymbol{\epsilon}}
\left[
\left\|
\boldsymbol{\epsilon}
-
\boldsymbol{\epsilon}_{\theta}
(\mathbf{A}^{k}_t,k,\mathbf{h}_t)
\right\|_2^2
\right].
\]

At inference, the action sequence is refined from Gaussian noise using 10 DDIM steps. SPICE executes the first \(T_a=8\) of the \(T_p=16\) predicted joint positions before replanning from the latest observations, allowing the motion to adapt as tactile interaction changes.
\subsection{Data collection and Experimental setup}

We used a controlled mock plant because real foliage is cluttered and dynamic, making tactile demonstrations sensitive to plant motion. As haptic feedback was not economically feasible, the operator relied on direct observation and a tactile heat map; the mock setup was therefore occluded for the robot but observable to the operator.
\begin{figure}[h]
    \centering
    \vspace{5pt}
    \includegraphics[width=.4\textwidth]{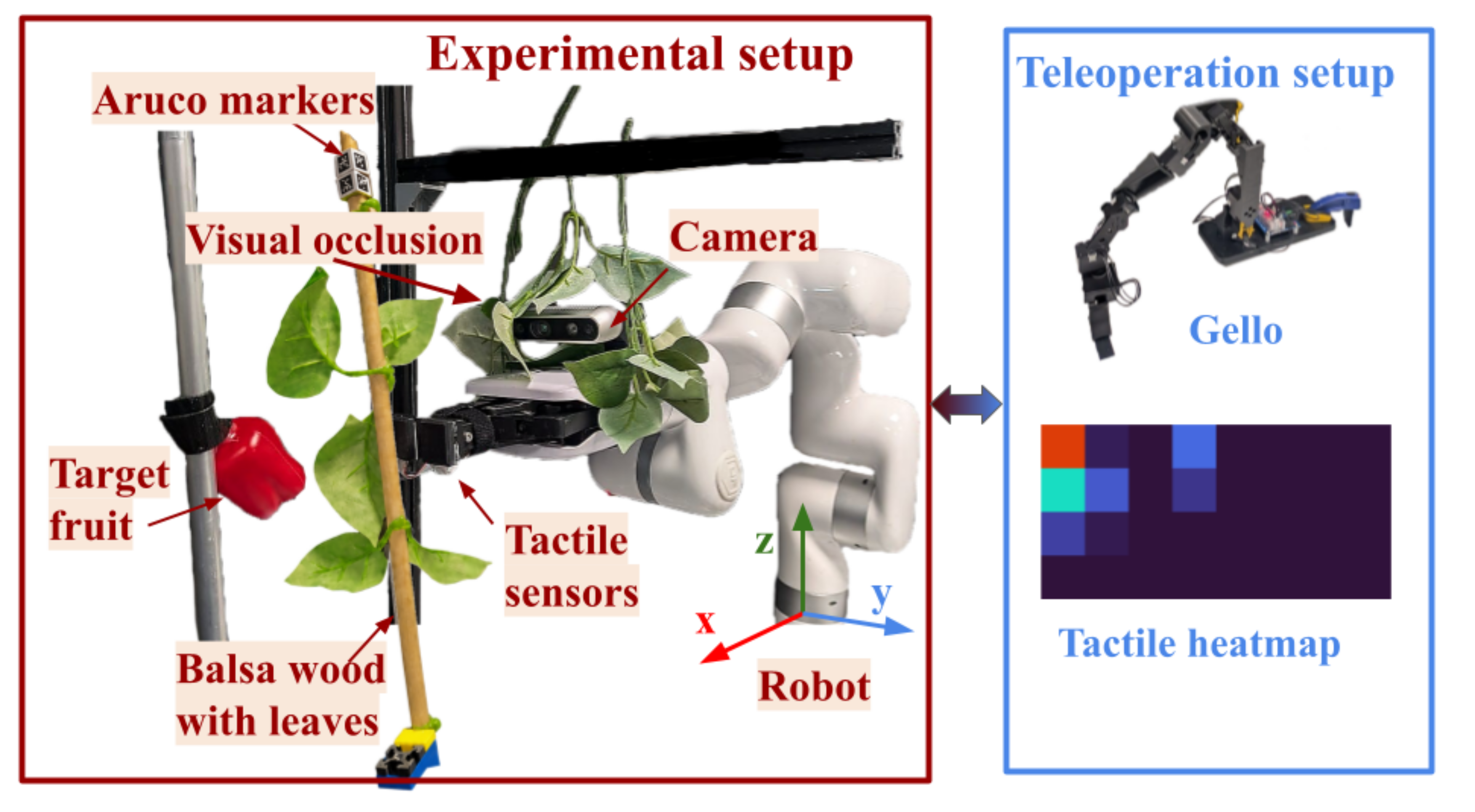}
        \vspace{-10pt}
    \caption{Mock branch setup for demonstration collection and evaluation under visual occlusion, and the GELLO teleoperation setup with tactile heat-map feedback.}
    \label{fig:demo_setup}
        \vspace{-15pt}
\end{figure}
The setup (Fig.~\ref{fig:demo_setup}) used a UFactory xArm6 with two Xela uSkin uSPa44 tactile arrays mounted on the gripper front to detect contact in clutter. An artificial red fruit served as the target, while a \(9\,\mathrm{mm}\)-diameter balsa branch and artificial leaves formed the canopy. Branch clamped length was varied to change stiffness and contact location, with all elements mounted on a slotted aluminium frame for repeatable repositioning.

\begin{figure}[h]
        \vspace{-5pt}
    \centering
    \includegraphics[width=.5\textwidth]{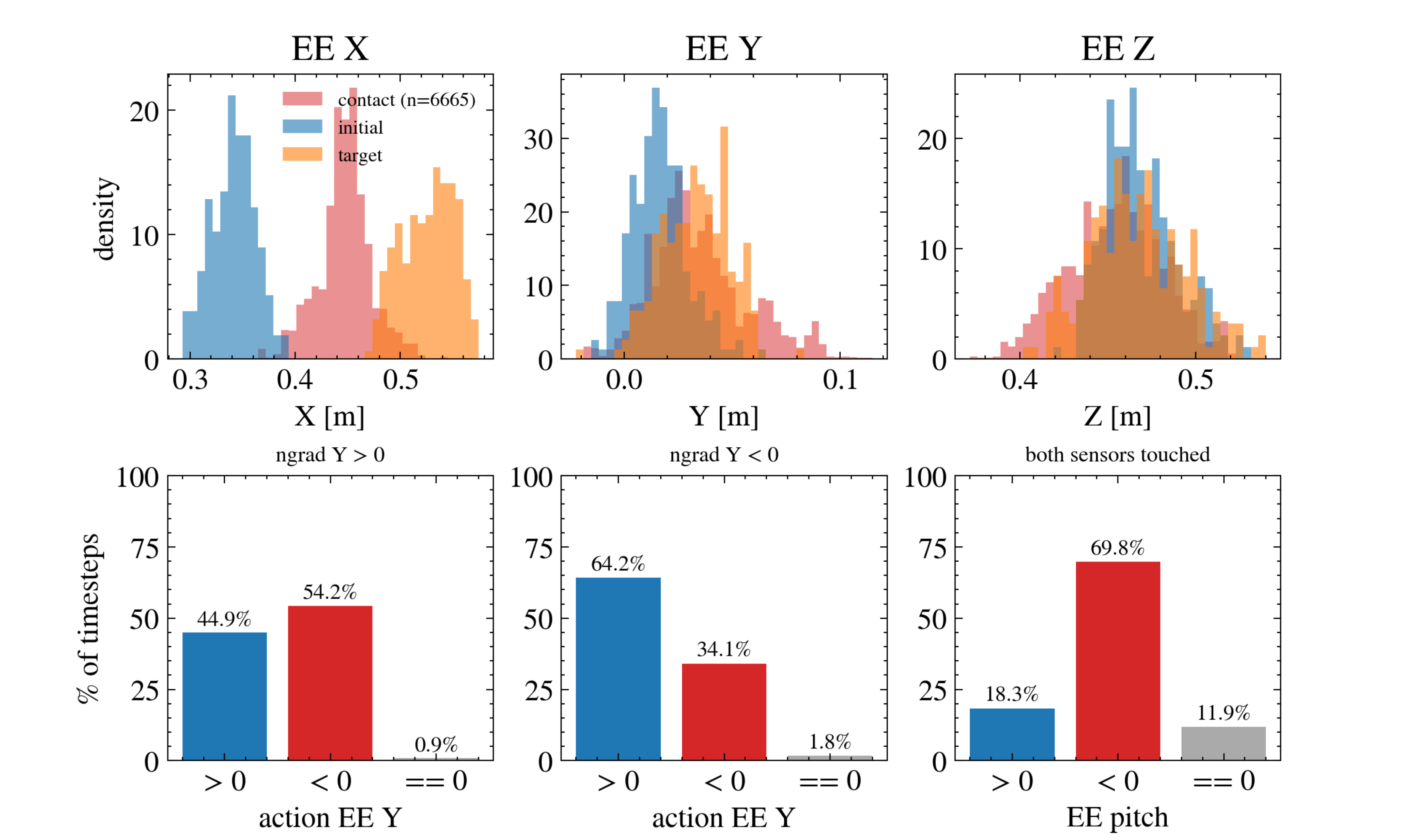}
        \vspace{-10pt}
\caption{ Distributions of end-effector positions and
interaction behaviours in the demonstration dataset.}
    \label{fig:data_distribution}
\end{figure}

Initial robot, target, and branch configurations were uniformly sampled. Initial end-effector positions were drawn from a \(10\times10\times10\,\mathrm{cm}\) workspace centred at \((0.30,\,0.01,\,0.45)\,\mathrm{m}\), with contact and target regions offset along the \(x\)-direction. The resulting distributions are shown in Fig.~\ref{fig:data_distribution}. A single operator collected demonstrations using GELLO~\cite{gello}, with direct observation and the tactile heat map. Data were recorded at \(30\,\mathrm{Hz}\); a wrist-mounted Intel RealSense camera was used only for the image-conditioned Diffusion Policy baseline.

We collected 240 demonstrations: 70 for each manoeuvring direction, 70 for reorientation and downward branch displacement, and 30 obstacle-free reaches. Manoeuvring directions were balanced across both sides of the branch. Because branch motion can alter the tactile gradient, failed, noisy, or incorrect demonstrations were removed. Opposite-sign lateral motion occurred in \(54.2\%\) and \(64.2\%\) of timesteps for positive and negative tactile gradients, respectively, while \(69.8\%\) of simultaneous two-sensor contact timesteps showed negative end-effector pitch (Fig.~\ref{fig:data_distribution}). Only 30 obstacle-free demonstrations were included because about \(70\%\) of contact-rich trajectories already contained free-space motion. Behaviour labels were not used during training.

Training used AdamW for 100 epochs with learning rate \(10^{-4}\), batch size 32, cosine decay with 500 warm-up steps, exponential moving averaging, and random seed 42.

\section{Evaluation Protocol and Metrics}
We evaluate SPICE using task success to determine whether the robot completes the required behaviour, together with interaction metrics that explain how the robot responds physically during contact.

\subsection{Experimental Protocol}
We evaluated target reaching under three sequential levels of interaction complexity: reaching the target without an obstructing branch, manoeuvring around a partially obstructing vertical branch, and manipulating a horizontal branch that blocks the target path and prevents lateral manoeuvring. Success was defined according to the physical response required in each task:
\begin{itemize}
    \item \textbf{Target reaching:} the final end-effector position is within \(3\,\mathrm{cm}\) of the target, corresponding to the diameter of the artificial fruit.
    \item \textbf{Branch manoeuvring:} the end effector moves laterally around the branch before reaching the target.
    \item \textbf{Branch manipulation:} the end effector reorients above the branch, displaces it downward by at least \(5\,\mathrm{cm}\) to create sufficient access to the target, and then reach the target.
\end{itemize}
A method advanced to the next task only if it achieved at least \(50\%\) success within the first 10 trials of the current task.

Across experiments, initial and target joint configurations were generated by independently sampling each joint from the demonstration distributions. A pool of 180 initial--target pairs was allocated across experiments, with branches placed at eight predefined locations and contacts balanced between the two tactile sensors.

Trials were terminated when continued execution risked damaging the branch or tactile sensors, including sustained loading without a contact-responsive change in motion, increasing compression towards the contacted side, or repeated motion without escaping contact for more than \(5\,\mathrm{s}\). Terminated trials were counted as unsuccessful. Because tactile sensing covered only the end-effector contact faces, branches moving beyond this region were manually released when necessary after the evaluated response. Contact metrics were computed before intervention.

\subsection{Experimental Metrics}
Task success determines whether the required behaviour is completed, but does not reveal how the robot achieves it. We therefore use metrics that correspond directly to the physical response required by each task: reaching accuracy for all tasks, lateral redirection and tactile loading for manoeuvring, and reorientation and branch displacement for manipulation.

\paragraph{General metrics}
All tasks ultimately require reaching the target. \textbf{Target error} therefore measures final reaching accuracy: $e_g =
\left\|
\mathbf{p}_{\mathrm{ee}}^{\mathrm{final}}-\mathbf{p}_g
\right\|_2 ,$ where \(\mathbf{p}_{\mathrm{ee}}^{\mathrm{final}}\) and \(\mathbf{p}_g\) are the final end-effector and target positions. Because SPICE replans action chunks during execution, we additionally measure \textbf{action-chunk discontinuity} to determine whether replanning introduces abrupt changes between consecutive trajectories: $d_k =
\left\|
\mathbf{q}_{k+1}^{\mathrm{start}}
-\mathbf{q}_k^{\mathrm{end}}
\right\|_2$. For each trial, \(d_k\) is averaged across replanning boundaries and reported in degrees; lower values indicate smoother transitions.

\paragraph{Branch manoeuvring}
Manoeuvring requires the robot to redirect its motion laterally rather than continue pushing into the branch. We therefore measure the direction of end-effector motion relative to the branch and the tactile loading produced during this response. \textbf{Manoeuvring ratio} measures lateral motion relative to forward motion during contact:$ R_{yx} =
\frac{\left|\Delta y_{\mathrm{ee}}\right|}
     {\left|\Delta x_{\mathrm{ee}}\right|}.$ Forward displacement represents motion into the branch, while lateral displacement represents motion around it. Larger \(R_{yx}\) therefore indicates a greater lateral component relative to forward pushing. To quantify sustained environmental interaction during manoeuvring, tactile magnitude is defined as $T_t = \max_i \left\|\mathbf{t}_{i,t}\right\|_2 .$ Because branch motion can produce transient peaks, \(T_t\) is smoothed using a centred \(2\,\mathrm{s}\) rolling average. \textbf{Peak tactile response} is the maximum of this smoothed signal during contact.

\paragraph{Branch manipulation}
Manipulation requires two coordinated responses: reorienting the end effector above the branch and then displacing the branch downward. We therefore quantify these two stages separately.

\textbf{End-effector pitch change} measures reorientation after contact:$\Delta\theta = \max_{t\in\mathcal{C}} \left|\theta_t-\theta_{t_c}\right|,$ where \(t_c\) denotes contact onset and \(\mathcal{C}\) the contact interval.

\textbf{Branch displacement} measures the resulting downward branch motion: $\Delta z_b = z_b(t_c)-\min_{t\in\mathcal{C}} z_b(t).$ Branch position is measured in the robot-base frame using an ArUco marker. Together, \(\Delta\theta\) and \(\Delta z_b\) determine whether the robot performs the coordinated reorientation and displacement required for branch manipulation rather than reaching the target through incidental contact.

\section{Experiments and results}
We evaluated SPICE through five experiments examining tactile representation, encoder design, reactive-control comparison, robustness under visual occlusion, and zero-shot transfer to real plants. We first assess its overall performance by aggregating 170 SPICE trials across these experiments, comprising 40 target-reaching, 90 branch-manoeuvring, and 40 branch-manipulation trials. SPICE achieved success rates of \(100\%\), \(98.9\%\), and \(85.0\%\), respectively. The following sections examine these results in detail.

\subsection{Experiment I: Tactile Representation}
Our first experiment evaluates how tactile representations motivated by prior work affect SPICE performance in target reaching and branch manoeuvring. We compare SPICE with mean three-axis tactile values (\(\mathrm{T}_{xyz}\)-DP), inspired by TacDiffusion~\cite{wu2025tacdiffusion}, and direct taxel input (TactDP), motivated by Reactive Diffusion Policy~\cite{xue2025reactive}. Because Reactive Diffusion Policy applies PCA to high-dimensional camera-based tactile observations, we use the raw taxel measurements directly. We also adapt the tactile-gradient formulation from RICE~\cite{rice} to form three-axis (RICE-DP) and \(y\)-axis (RICE-y-DP) representations, with the latter aligned to the lateral manoeuvring direction.
SPICE, TactDP, \(\mathrm{T}_{xyz}\)-DP, and RICE-DP were first evaluated on the no-branch task, where all achieved \(10/10\) target-reaching successes.
In branch manoeuvring, SPICE achieved \(19/20\) successes (\(95\%\)), substantially outperforming all alternatives. RICE-y-DP was next best with \(8/20\) (\(40\%\)), followed by TactDP without \(c_t\) (\(7/20\)), TactDP with \(c_t\) (\(5/20\)), and RICE-DP (\(3/20\)).
To assess the contribution of \(c_t\), SPICE, \(\mathrm{T}_{xyz}\)-DP, and TactDP were tested with and without contact conditioning. Both \(\mathrm{T}_{xyz}\)-DP variants and SPICE without \(c_t\) failed all 10 initial trials. Without \(c_t\), \(\mathrm{T}_{xyz}\)-DP continued pushing after contact; with \(c_t\), it generated lateral motion but failed to clear the branch. TactDP without \(c_t\) occasionally succeeded through irregular motions, whereas with \(c_t\) it produced more consistent contact-directed behaviour.
\begin{figure}[h!]
    \centering
    \includegraphics[width=.8\columnwidth]{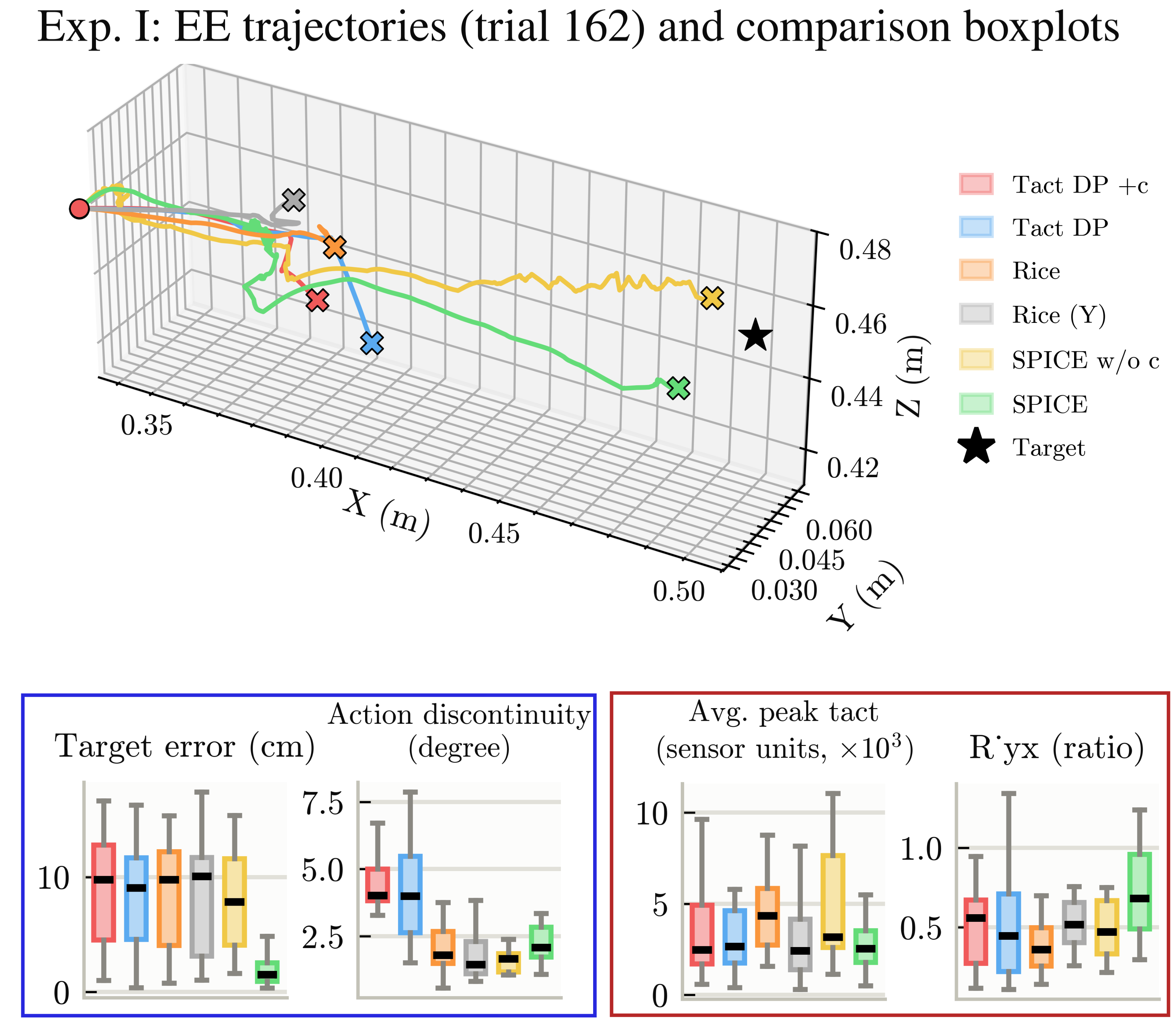}
    \caption{Comparison of End Effector (EE) trajectories of SPICE with aggregated, direct-taxel, and
    hand-crafted tactile representations in vertical-branch trials.
    Representative trajectories show contact responses from a matched
    initial condition. Box plots report target error, action
    discontinuity, Peak Tactile, and \(R_{yx}\).}
    \label{fig:tac_rep}
\end{figure}
SPICE also achieved the lowest mean target error (\(1.73\pm1.11\,\mathrm{cm}\)), highest \(R_{yx}\) (\(0.69\pm0.30\)), and lowest mean peak tactile response (\(2766\pm1344\) sensor units). Results are summarised in Table~\ref{tab:tact_rep} and Fig.~\ref{fig:tac_rep}.
These results suggest that local spatial tactile structure and explicit contact conditioning both support reliable branch manoeuvring. The stronger performance of RICE-y-DP indicates that behaviour-aligned tactile gradients provide useful directional cues, while adding \(x\)- and \(z\)-axis gradients did not improve performance. Direct taxel conditioning retained the full measurements but produced less consistent responses. Because all alternatives remained below the \(50\%\) progression criterion, they were not evaluated on branch manipulation.

\begin{table}
\vspace{15pt}
\centering
\caption{Expt.~I. Tactile representation}
\label{tab:tact_rep}
\scriptsize
\setlength{\tabcolsep}{1.5pt}
\renewcommand{\arraystretch}{1.12}

\resizebox{\columnwidth}{!}{%
\begin{tabular}{lccccc}
\toprule
\textbf{Policy} &
\shortstack{\textbf{Task}\\\textbf{success}} &
\shortstack{\textbf{Target error}\\\textbf{(cm)}} &
\shortstack{\textbf{Discontinuity}\\\textbf{($^\circ$/replan)}} &
\shortstack{\textbf{Peak Tactile}\\\textbf{(sensor units)}} &
\shortstack{\(\boldsymbol{R_{yx}}\)\\\textbf{during contact}} \\
\midrule

\(\mathrm{F}_{xyz}\)-DP \(+c_t\)
& 0/10 (0\%)
& $6.68 \pm 3.69$
& $3.77 \pm 1.84$
& $\mathbf{2097 \pm 2300.42}$
& $0.62 \pm 0.25$ \\

$\mathrm{F}_{xyz}$-DP w/o $c_t$
& 0/10 (0\%)
& $8.92 \pm 0.31$
& $5.83 \pm 1.54$
& $7150.11 \pm 932.88$
& $0.32 \pm 0.20$ \\

RICE-y-DP $+c_t$
& 8/20 (40\%)
& $8.20 \pm 5.12$
& $1.93 \pm 1.11$
& $3704.65 \pm 3509.58$
& $0.62 \pm 0.49$ \\

RICE-DP $+c_t$
& 3/20 (15\%)
& $8.64 \pm 4.81$
& $2.08 \pm 0.85$
& $4574.69 \pm 2098.38$
& $0.44 \pm 0.26$ \\

TactDP w/o $c_t$
& 7/20 (35\%)
& $8.40 \pm 4.39$
& $4.08 \pm 1.87$
& $3853.03 \pm 3397.36$
& $0.58 \pm 0.46$ \\

SPICE w/o $c_t$
& 0/10 (0\%)
& $8.48 \pm 4.89$
& $\mathbf{1.62 \pm 0.42}$
& $4909.29 \pm 3382.96$
& $0.48 \pm 0.18$ \\

SPICE
& \textbf{19/20 (95\%)}
& $\mathbf{1.73 \pm 1.11}$
& $\mathbf{2.31 \pm 0.85}$
& $\mathbf{2766.28 \pm 1343.60}$
& $\mathbf{0.69 \pm 0.30}$ \\

\bottomrule
\end{tabular}%
}
\vspace{-15pt}
\end{table}

\subsection{Experiment II: SPICE encoder comparison}

\begin{figure}[h!]
    \centering
\includegraphics[width=0.4\textwidth,trim={0cm 0.cm 0 0},clip]{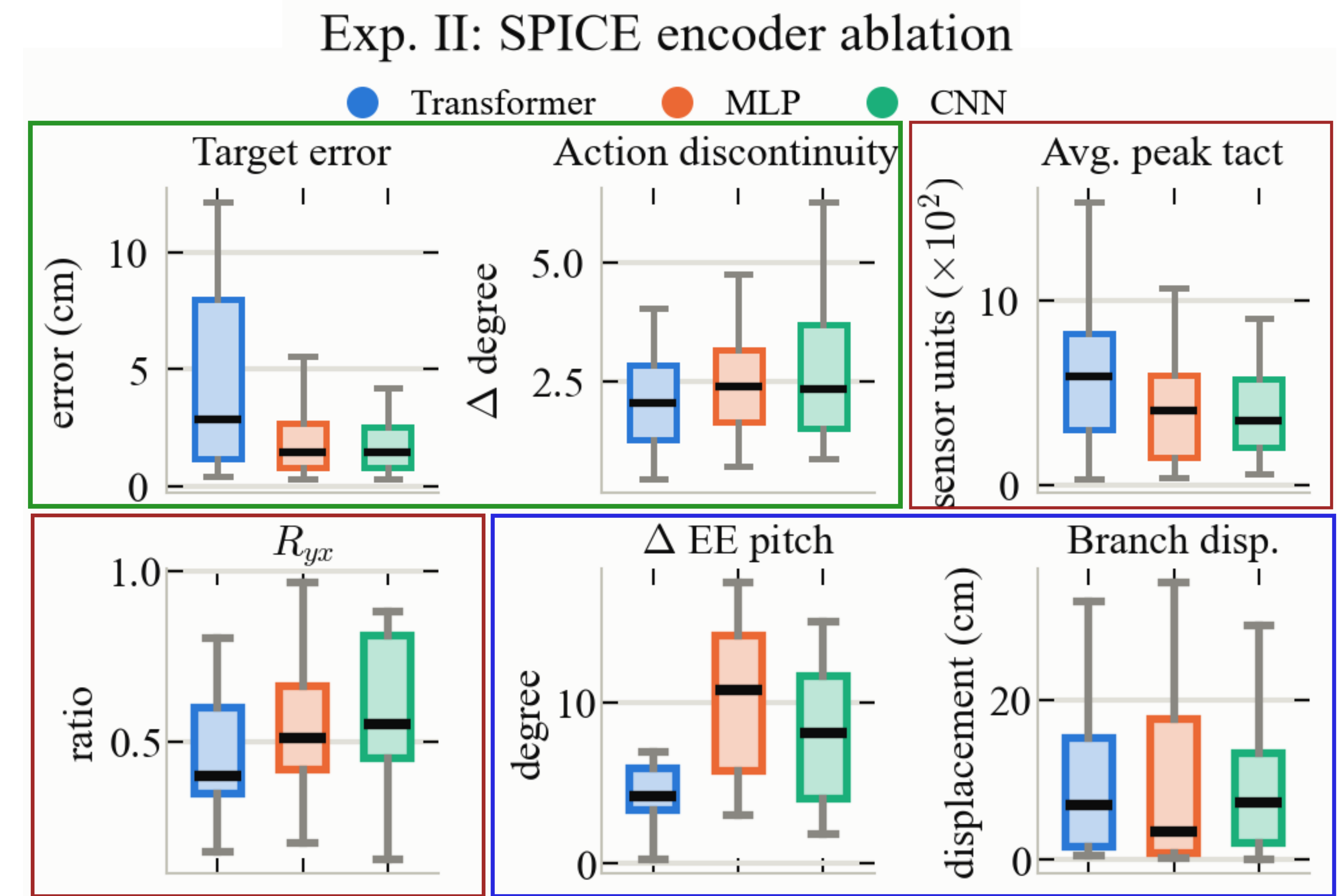}
     \vspace{-5pt}
    \caption{Tactile-encoder ablation comparing MLP, Transformer, and CNN encoders. Box plots report target error and peak sustained tactile response for horizontal and vertical trials, action-chunk discontinuity during contact, the vertical lateral-to-forward displacement ratio, and horizontal end-effector pitch change and branch displacement. Lower discontinuity indicates smoother transitions between action chunks.}
    \label{fig:ablation}
    \vspace{-5pt}
\end{figure}

We next compared SPICE variants using CNN, MLP, and Transformer tactile encoders to examine how encoder architecture affects contact-responsive behaviour. The MLP uses flattened taxel measurements without explicit spatial locality, while the Transformer uses positional embeddings and self-attention to model relationships across taxels. All variants used the same dataset, embedding dimension, diffusion backbone, training procedure, and test conditions.
All encoders achieved \(10/10\) successes in no-branch reaching. With branch interaction, CNN and MLP both achieved \(20/20\) manoeuvring successes, while Transformer achieved \(14/20\). CNN and MLP showed similar target accuracy, \(R_{yx}\), and peak tactile response (Fig. \ref{fig:ablation}). Transformer had larger target errors, lower median \(R_{yx}\), and higher peak tactile response, indicating less lateral motion and greater environmental loading. In unsuccessful trials, insufficient lateral motion led to termination despite smooth action transitions.
In branch manipulation, CNN achieved \(18/20\) successes, including four recoveries through retreat and renewed contact, compared with \(11/20\) for MLP and \(1/20\) for Transformer. The two unsuccessful CNN trials moved above the branch without displacing it. MLP showed similar pitch reorientation but lower branch displacement, with failures involving post-contact oscillations and slips. Transformer frequently produced repeated vertical and lateral motions without completing the task. Overall, CNN most consistently produced the coordinated reorientation and downward displacement required for manipulation.
These results suggest that the CNN's local spatial representation better supports increasingly complex contact responses and recovery. CNN was therefore selected as SPICE's tactile encoder.

\subsection{Experiment III: Comparison with RICE}
\begin{figure*}[t]
\includegraphics[width=.95\textwidth,angle=-0]{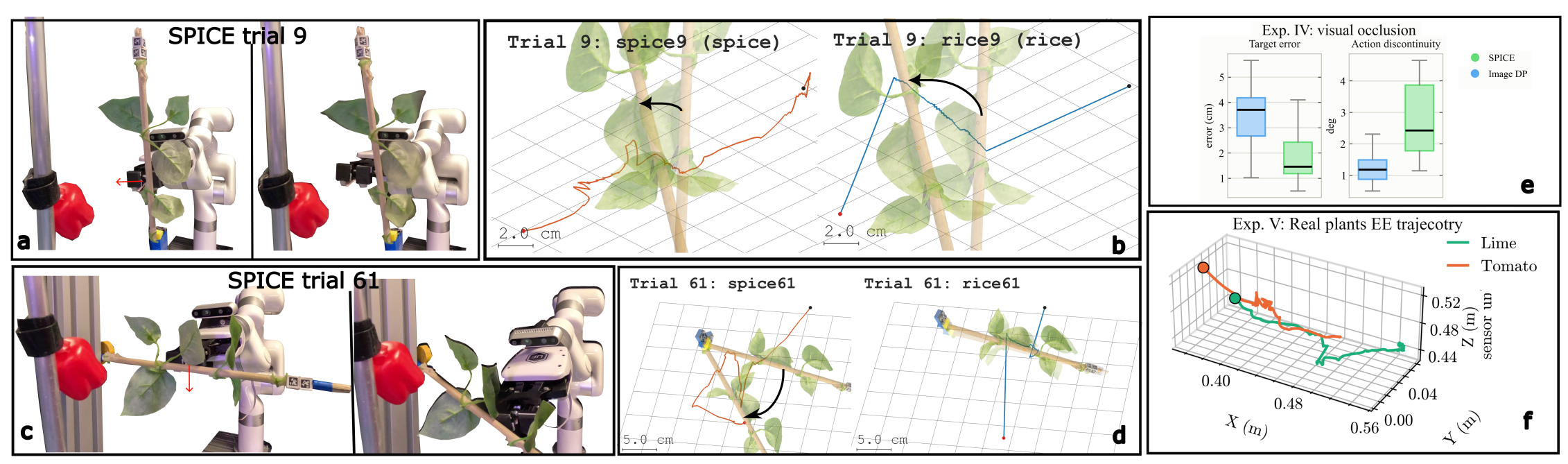}
    \caption{comparison of SPICE with RICE. (a) Robot configuration before and after contact in vertical-branch trial 9, where SPICE manoeuvres around the branch. (b) End-effector and branch trajectories for SPICE and RICE in trial 9. SPICE moves around the branch, whereas RICE initially moves along it before passing around it. (c) Robot configuration before and after contact in horizontal-branch trial 61, where SPICE pushes the branch downward before proceeding toward the target. (d) End-effector and branch trajectories for SPICE and RICE in trial 61. SPICE displaces the branch from its path, whereas RICE does not. (e) Box plot comparison of ImageDP and SPICE. (f) EE trajectory lime and tomato real plant experiment.}
    \vspace{-10pt}
    \label{fig:rice_spice_trajectories}
\end{figure*}

We compared SPICE with the reactive RICE controller~\cite{rice}, which addresses the same cluttered-reaching objective using predefined target and tactile-interaction costs. Both methods used the same robot setup, initial--target configurations, and branch poses across 30 manoeuvring and 20 manipulation trials. To evaluate oblique contact, branch orientation was varied by \(-10^\circ\), \(-8^\circ\), \(0^\circ\), \(8^\circ\), and \(10^\circ\) about the robot-base \(y\)-axis. Interaction behaviour was also compared using RICE's cost function, reconstructed for SPICE from recorded tactile measurements, end-effector poses, and taxel locations.
\begin{figure}[h!]
    \centering
    \includegraphics[width=.8\columnwidth]{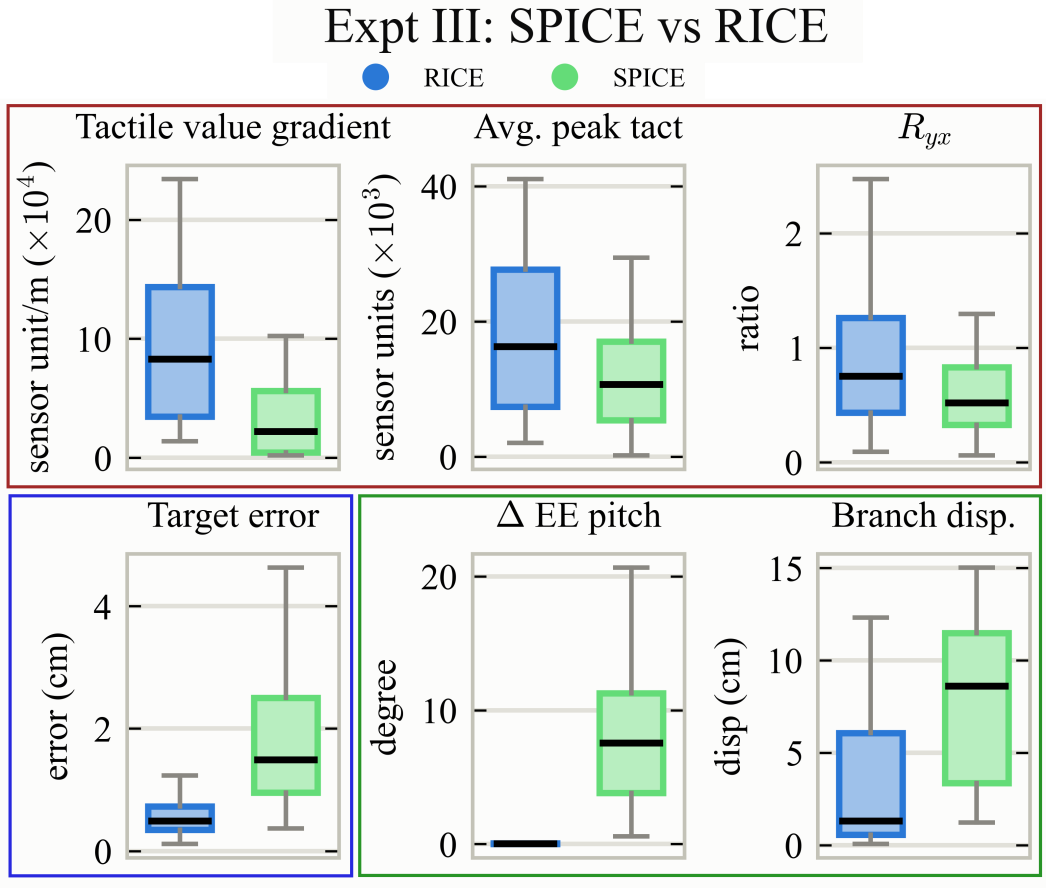}
    \caption{Comparison of SPICE and RICE. Box plots show target error,
    vertical tactile-gradient magnitude, peak sustained tactile response,
    \(R_{yx}\), horizontal pitch change, and downward branch
    displacement. RICE followed oblique vertical branches, whereas SPICE
    reduced contact and displaced horizontal branches from the target
    path.}
    \label{fig:rice_spice}
    \vspace{-5pt}
\end{figure}
No-branch reaching was omitted because both methods had already demonstrated this capability. Both achieved \(30/30\) manoeuvring successes. SPICE more often reduced or released contact before resuming target-directed motion, with lower tactile-gradient magnitude (\((3.32\pm3.27)\times10^{4}\) versus \((10.03\pm7.73)\times10^{4}\) sensor unit/m) and peak tactile response (\((12.72\pm11.64)\times10^{3}\) versus \((18.53\pm13.53)\times10^{3}\) sensor units). Fig~\ref{fig:rice_spice} shows the comparison.  RICE instead tended to follow the angled branch while maintaining contact (Fig \ref{fig:rice_spice_trajectories}), with larger \(R_{yx}\) (\(0.87\pm0.54\) versus \(0.67\pm0.52\)) reflecting greater lateral motion along the branch.

\begin{table}[h!]
\centering
\caption{Exp III: SPICE vs RICE }
\label{tab:spice_rice}
\resizebox{\columnwidth}{!}{%
\begin{tabular}{llcc}
\toprule
\textbf{Task} & \textbf{Metric} & \textbf{RICE} & \textbf{SPICE} \\
\midrule
\multirow{4}{*}{\begin{tabular}[c]{@{}l@{}}Branch\\manoeuvring\end{tabular}}
& Task success
& \textbf{30/30 (100\%)} 
& \textbf{30/30 (100\%)} \\
& Tactile gradient (sensor unit/m, $\times 10^{4}$)
& $10.03 \pm 7.73$ 
& $\mathbf{3.32 \pm 3.27}$\\

& Peak tactile response (sensor units, $\times 10^{3}$)
& $18.53 \pm 13.53$ 
& $\mathbf{12.72 \pm 11.64}$ \\

& $R_{yx}$
& $0.87 \pm 0.54$ 
& $\mathbf{0.67 \pm 0.52}$ \\
\midrule
\multirow{3}{*}{\begin{tabular}[c]{@{}l@{}}Branch\\manipulation\end{tabular}}
& Task success
& 0/30 (0\%) 
& \textbf{20/20 (100\%)} \\

& $\Delta$ EE pitch ($^\circ$)
& $0.05 \pm 0.05$
& $\mathbf{7.95 \pm 5.45}$ \\

& Branch displacement (cm)
& $5.74 \pm 8.89$ 
& $\mathbf{9.02 \pm 7.40}$  \\

\midrule
\multirow{1}{*}{\begin{tabular}[c]{@{}l@{}}Reach target\end{tabular}}
& Target error (cm)
& $\mathbf{1.17 \pm 2.48}$
& $2.78 \pm 5.22$ \\
\bottomrule
\end{tabular}%
}
\vspace{-15pt}
\end{table}

In branch manipulation, SPICE achieved \(20/20\) successes, whereas RICE did not satisfy the manipulation criterion. SPICE produced greater end-effector pitch change (\(7.95\pm5.45^\circ\) versus \(0.05\pm0.05^\circ\)) and branch displacement (\(9.02\pm7.40\) versus \(5.74\pm8.89\,\mathrm{cm}\)), reflecting the coordinated reorientation and downward motion required for manipulation. RICE primarily displaced the branch laterally but achieved lower target error (\(1.17\pm2.48\) versus \(2.78\pm5.22\,\mathrm{cm}\); Table~\ref{tab:spice_rice}).
Overall, SPICE added multi-stage branch manipulation while retaining target-reaching and manoeuvring capabilities. RICE achieved higher target accuracy but tended to maintain longer contact under oblique branch orientations.
\subsection{Experiment IV: Visual occlusion}
\begin{table}[h!]
 \vspace{-10pt}
\centering
\caption{Exp. IV: Visual occlusion (SPICE vs. Image DP)}
\label{tab:image_spice}
\vspace{-8pt}
\begin{tabular}{lcc}
\toprule
\textbf{Metric} & \textbf{Image DP} & \textbf{SPICE} \\
\midrule
Target error (cm)
& $3.44 \pm 1.43$
& $\mathbf{1.97 \pm 1.28}$ \\

Action discontinuity ($^\circ$)
& $\mathbf{1.24 \pm 0.52}$
& $2.77 \pm 1.12$ \\

Reach target success (no branch) 
& $13/20$
& $\mathbf{19/20}$ \\

Branch manoeuvre task success
& $0/20$ 
& $\mathbf{18/20}$  \\
\bottomrule
\end{tabular}%
\vspace{-10pt}
\end{table}

To assess SPICE's robustness to visual occlusion, we compared it with an image-conditioned Diffusion Policy trained from the same demonstrations, replacing tactile input with wrist-camera images. Evaluation used the same visually occluded conditions present during demonstration collection. In target reaching, SPICE achieved \(19/20\) successes compared with \(13/20\) for Image DP, with lower mean target error (\(1.97\pm1.28\) versus \(3.44\pm1.43\,\mathrm{cm}\)) (Fig~\ref{tab:image_spice}). Image DP produced smoother action transitions (\(1.24\pm0.52^\circ\) versus \(2.77\pm1.12^\circ\)), but lower task success as shown in Fig~\ref{fig:rice_spice_trajectories}(e). With branch interaction, SPICE achieved \(18/20\) manoeuvring successes, whereas Image DP achieved \(0/20\) and was not progressed to branch manipulation. Tactile-response metrics were not compared because they were not recorded for Image DP.
These results indicate that SPICE's tactile conditioning supported both target reaching and contact-responsive behaviour under visual occlusion, whereas image conditioning was less reliable for reaching and did not produce effective contact responses.

\subsection{Experiment V: Zero-Shot Transfer to Real Plants}
\begin{figure}[h!]
    \centering
    \includegraphics[width=.4\textwidth]{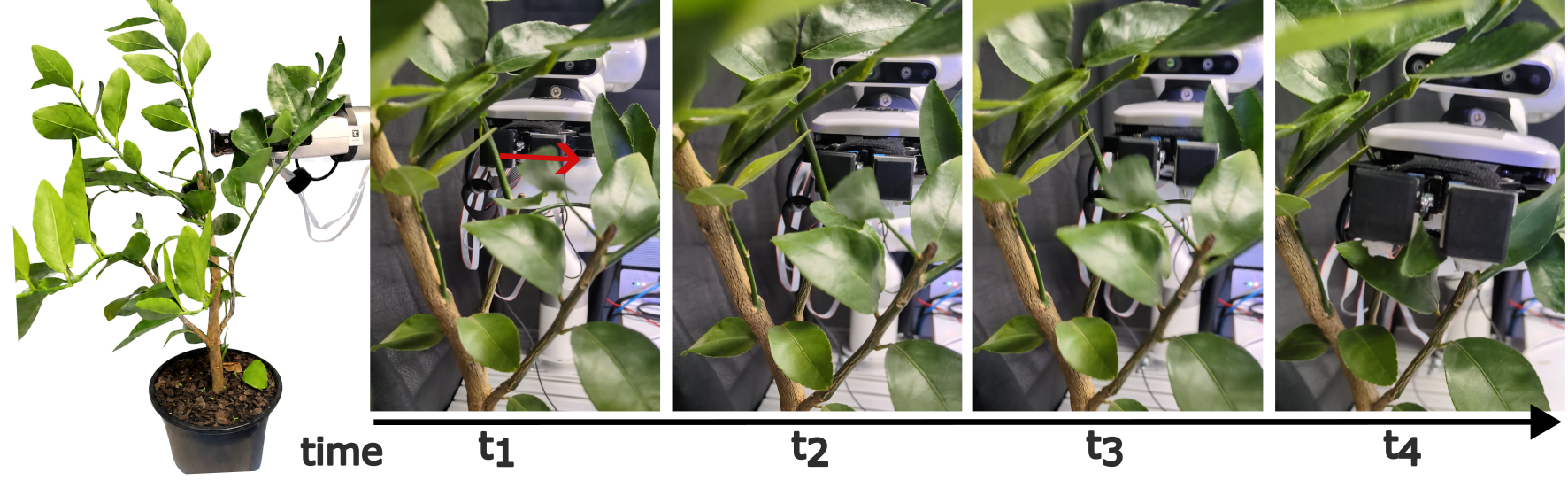}
    \caption{Image sequence of SPICE manoeuvring around a lime branch.}
    \label{fig:lime}
\end{figure}

\begin{figure}[h!]
    \centering
    \includegraphics[width=.4\textwidth]{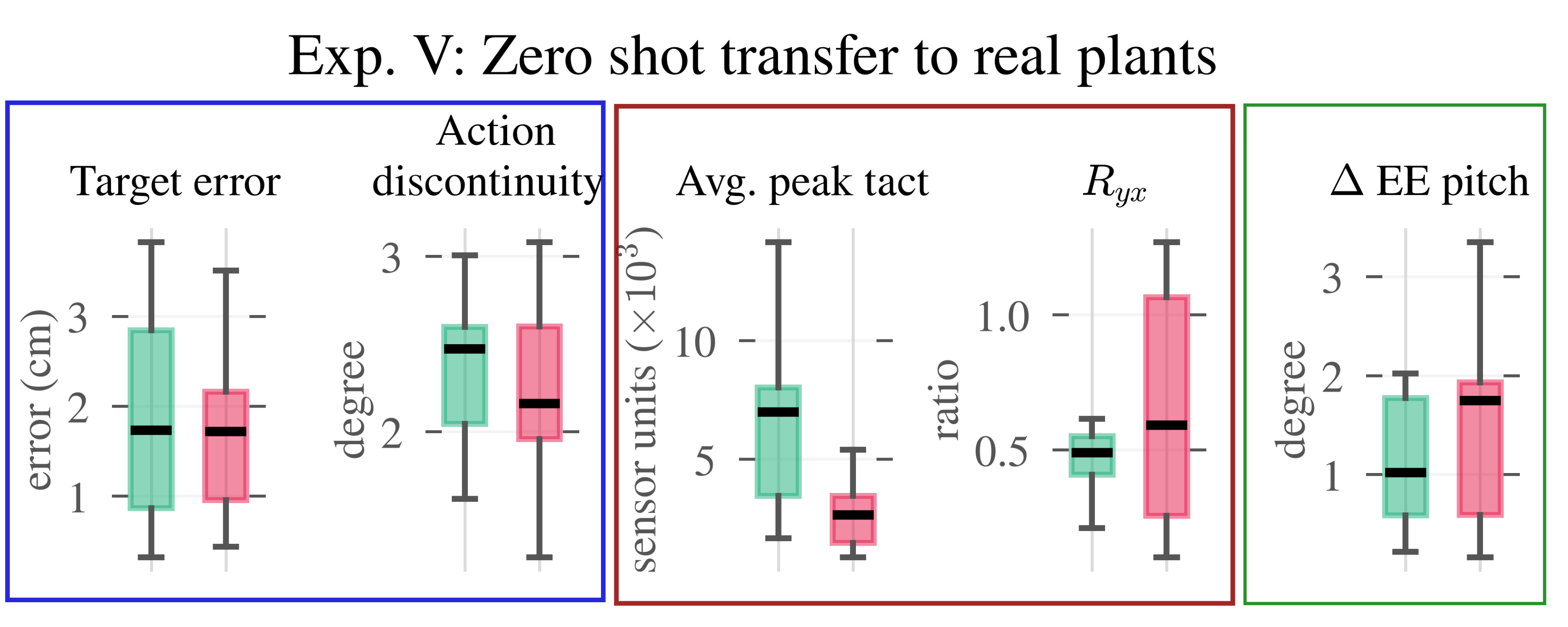}
    \vspace{-5pt}
    \caption{Box plot comparison for lime and tomato experiments}
    \vspace{-10pt}
    \label{fig:real_plants}
\end{figure}

We evaluated SPICE for zero-shot transfer to previously unseen real plants. Lime (Fig~\ref{fig:lime}) and tomato (Fig~\ref{fig:tomato_fig}) plants differed from the demonstration setup in branch structure, compliance, and foliage dynamics. Across 40 trials, SPICE successfully reached the target in all 20 lime trials and 15 of 20 tomato trials; the remaining five tomato trials were terminated by safety stops. Plant pots were repositioned between trials to vary branch pose, contact location, and approach geometry.
Figure~\ref{fig:real_plants} shows similar median target errors for both species (approximately \(1.8\,\mathrm{cm}\)) but different interaction behaviours. Lime trials showed a higher median peak tactile response (approximately \(7\times10^3\) versus \(2.7\times10^3\) sensor units), consistent with interaction with thinner, stiffer branches, which SPICE generally manoeuvred around, including horizontal branches. Tomato trials showed a larger median end-effector pitch change (approximately \(1.8^\circ\) versus \(1.0^\circ\)) and a broader \(R_{yx}\) distribution. In the denser, more compliant foliage, SPICE more often displaced horizontal branches, manoeuvred around vertical stems, and generated lateral or retreat corrections as the foliage moved.
Overall, SPICE transferred to both unseen plant species without fine-tuning and adapted its interaction behaviour to their different structures. The five safety stops in tomato trials highlight the limitations of end-effector-only tactile sensing in dense foliage.

\section{Discussion}

The experiments validate SPICE as a single tactile-conditioned policy for target reaching, branch manoeuvring, and branch manipulation under visual occlusion, with \(87.5\%\) zero-shot success on unseen real plants. It outperformed the evaluated tactile and image-conditioned baselines, achieving \(95\%\) branch-manoeuvring success versus \(40\%\) for the strongest tactile representation, while extending beyond RICE to branch manipulation.
The tactile ablations highlight the importance of spatial structure. Aggregated forces can remain high as contact shrinks, whereas gradients capture changes in contact distribution. RICE-y-DP outperforming RICE-DP suggests that behaviour-aligned cues are particularly useful, while the CNN’s stronger manipulation and recovery performance indicates the value of richer local taxel relationships.
RICE achieved accurate reaching using instantaneous tactile-gradient direction but did not regulate interaction magnitude, leading to prolonged contact for misaligned branches. SPICE instead adapted through contact reduction, retreat, and reorientation, supporting learned policies when the appropriate response changes during interaction.
SPICE also produced behaviours not directly demonstrated, including retreat, re-contact, and alternative manoeuvring on unseen plants, suggesting adaptation beyond trajectory reproduction. Limitations include single-operator, non-haptic demonstrations and end-effector-only tactile coverage. Future work will explore richer demonstrations, broader tactile sensing, and visuotactile policies combining global visual guidance with local tactile adaptation.

\section{Conclusion}

This work presented SPICE, a tactile-conditioned diffusion policy for reaching through visually occluded plant canopies. From demonstrations collected using a simple, cost-efficient setup, SPICE learned three contact-responsive behaviours: target reaching, branch manoeuvring, and branch manipulation, despite intermittent and unevenly represented contact interactions. Across more than 180 real-robot trials, SPICE was evaluated against alternative tactile representations, encoder architectures, reactive control, and image-conditioned learning. It further demonstrated zero-shot transfer across 40 trials on lime and tomato plants with different branch structures and interaction dynamics. To the best of the authors' knowledge, SPICE is the first tactile-conditioned diffusion policy for reaching through physically obstructed plant canopies, demonstrating that tactile-conditioned generative policies can support diverse physical interactions under severe visual occlusion.









\bibliographystyle{IEEEtran}
\bibliography{reference}

@inproceedings{chi2023diffusion,
  author    = {Cheng Chi and Zhenjia Xu and Siyuan Feng and
               Eric Cousineau and Yilun Du and Benjamin Burchfiel and
               Russ Tedrake and Shuran Song},
  title     = {Diffusion Policy: Visuomotor Policy Learning via Action Diffusion},
  booktitle = {RSS},
  year      = {2023},
}

@article{barreiros2025careful,
  author  = {{TRI LBM Team} and Jose Barreiros and Andrew Beaulieu and
             Aditya Bhat and Rick Cory and Eric Cousineau and Hongkai Dai and
             Ching-Hsin Fang and Kunimatsu Hashimoto anders},
  title   = {A Careful Examination of Large Behavior Models f
             Muhammad Zubair Irshad and Masha Itkina and othor
             Multitask Dexterous Manipulation},
  journal = {arXiv preprint arXiv:2507.05331},
  year    = {2025},
  doi     = {10.48550/arXiv.2507.05331}
}

@inproceedings{zhao2025polytouch,
  author    = {Jialiang Zhao and Naveen Kuppuswamy and Siyuan Feng and
               Benjamin Burchfiel and Edward Adelson},
  title     = {{PolyTouch}: A Robust Multi-Modal Tactile Sensor for
               Contact-Rich Manipulation Using Tactile-Diffusion Policies},
  booktitle = {ICRA},
  year      = {2025}
}

@inproceedings{rice,
  author    = {Nidhi Homey Parayil and  Thierry Peynot and Chris Lehnert},
  title     = {RICE: Reactive Interaction Controller for Cluttered Canopy Environment},
  booktitle = {ICRA)},
  year      = {2026}
}

@inproceedings{xue2025reactive,
  author    = {Han Xue and Jieji Ren and Wendi Chen and Gu Zhang and
               Yuan Fang and Guoying Gu and Huazhe Xu and Cewu Lu},
  title     = {Reactive Diffusion Policy: Slow-Fast Visual-Tactile Policy
               Learning for Contact-Rich Manipulation},
  booktitle = {RSS},
  year      = {2025}
}

@inproceedings{jain2025learning,
  title={Learning to prune branches in modern tree-fruit orchards},
  author={Jain, Abhinav and Grimm, Cindy and Lee, Stefan},
  booktitle={ICRA},
  year={2025},
  organization={IEEE}
}

@misc{gello,
    title={GELLO: A General, Low-Cost, and Intuitive Teleoperation Framework for Robot Manipulators},
    author={Philipp Wu and Yide Shentu and Zhongke Yi and Xingyu Lin and Pieter Abbeel},
    year={2023},
}

@article{killpack2016model,
  title={Model predictive control for fast reaching in clutter},
  author={Killpack, Marc D and Kapusta, Ariel and Kemp, Charles C},
  journal={Autonomous Robots},
  volume={40},
  number={3},
  pages={537--560},
  year={2016},
  publisher={Springer}
}

@inproceedings{wu2025tacdiffusion,
  title={Tacdiffusion: Force-domain diffusion policy for precise tactile manipulation},
  author={Wu, Yansong and Chen, Zongxie and Wu, Fan and Chen, Lingyun and Zhang, Liding and Bing, Zhenshan and Swikir, Abdalla and Haddadin, Sami and Knoll, Alois},
  booktitle={ICRA},
  year={2025},
  organization={IEEE}
}

@article{helmut2025tactile,
  title={Tactile-conditioned diffusion policy for force-aware robotic manipulation},
  author={Helmut, Erik and Funk, Niklas and Schneider, Tim and de Farias, Cristiana and Peters, Jan},
  journal={arXiv preprint arXiv:2510.13324},
  year={2025}
}

@article{ablett2024multimodal,
  title={Multimodal and force-matched imitation learning with a see-through visuotactile sensor},
  author={Ablett, Trevor and Limoyo, Oliver and Sigal, Adam and Jilani, Affan and Kelly, Jonathan and Siddiqi, Kaleem and Hogan, Francois and Dudek, Gregory},
  journal={IEEE Transactions on Robotics},
  volume={41},
  year={2024},
  publisher={IEEE}
}

@article{yu2023mimictouch,
  title={Mimictouch: Leveraging multi-modal human tactile demonstrations for contact-rich manipulation},
  author={Yu, Kelin and Han, Yunhai and Wang, Qixian and Saxena, Vaibhav and Xu, Danfei and Zhao, Ye},
  journal={arXiv preprint arXiv:2310.16917},
  year={2023}
}

@article{wistreich2025dexskin,
  title={Dexskin: High-coverage conformable robotic skin for learning contact-rich manipulation},
  author={Wistreich, Suzannah and Shi, Baiyu and Tian, Stephen and Clarke, Samuel and Nath, Michael and Xu, Chengyi and Bao, Zhenan and Wu, Jiajun},
  journal={arXiv preprint arXiv:2509.18830},
  year={2025}
}

@article{xue2026tube,
  title={Tube Diffusion Policy: Reactive Visual-Tactile Policy Learning for Contact-rich Manipulation},
  author={Xue, Teng and Rigo, Alberto and Huang, Bingjian and Shen, Jiayi and Xu, Zhengtong and Colonnese, Nick and Memar, Amirhossein H},
  journal={arXiv preprint arXiv:2604.23609},
  year={2026}
}

@inproceedings{killpack2013fast,
  title={Fast reaching in clutter while regulating forces using model predictive control},
  author={Killpack, Marc D and Kemp, Charles C},
  booktitle={IEEE-RAS Humanoids)},
  year={2013},
  organization={IEEE}
}

@inproceedings{brouwer2024tactile,
  title={Tactile-informed action primitives mitigate jamming in dense clutter},
  author={Brouwer, Dane and Citron, Joshua and Choi, Hojung and Lepert, Marion and Lin, Michael and Bohg, Jeannette and Cutkosky, Mark},
  booktitle={ICRA},
  year={2024},
  organization={IEEE}
}

@article{nazari2023deep,
  title={Deep functional predictive control for strawberry cluster manipulation using tactile prediction},
  author={Nazari, Kiyanoush and Gandolfi, Gabriele and Talebpour, Zeynab and Rajendran, Vishnu and Rocco, Paolo and others},
  journal={arXiv preprint arXiv:2303.05393},
  year={2023}
}

@misc{rijal2026modelingbranchesactivemanipulation,
      title={Modeling Branches for Active Manipulation using Iterative Parameter Estimation}, 
      author={Madhav Rijal and Rashik Shrestha and Trevor Smith and Yu Gu},
      year={2026},
      eprint={2606.19314},
      archivePrefix={arXiv},
      primaryClass={cs.RO},
      url={https://arxiv.org/abs/2606.19314}, 
}

@article{fu2026push,
  title={Push-or-avoid: Deep reinforcement learning of obstacle-aware harvesting for orchard robots},
  author={Fu, Heng and Li, Tao and Feng, Qingchun and Chen, Liping},
  journal={Agriculture},
  volume={16},
  number={6},
  year={2026},
  publisher={MDPI}
}

@inproceedings{gursoy2024occlusion,
  title={Occlusion handling by pushing for enhanced fruit detection},
  author={Gursoy, Ege and Kuli{\'c}, Dana and Cherubini, Andrea},
  booktitle={IEEE/RSJ IROS},
  pages={76--81},
  year={2024},
  organization={IEEE}
}

@article{zhou2023branch,
  title={Branch interference sensing and handling by tactile enabled robotic apple harvesting},
  author={Zhou, Hongyu and Kang, Hanwen and Wang, Xing and Au, Wesley and Wang, Michael Yu and Chen, Chao},
  journal={Agronomy},
  volume={13},
  number={2},
  pages={503},
  year={2023},
  publisher={MDPI}
}

@inproceedings{you2022precision,
  title={Precision fruit tree pruning using a learned hybrid vision/interaction controller},
  author={You, Alexander and Kolano, Hannah and Parayil, Nidhi and Grimm, Cindy and Davidson, Joseph R},
  booktitle={ICRA},
  year={2022},
  organization={IEEE}
}

@inproceedings{yao2025safe,
  title={Safe leaf manipulation for accurate shape and pose estimation of occluded fruits},
  author={Yao, Shaoxiong and Pan, Sicong and Bennewitz, Maren and Hauser, Kris},
  booktitle={ICRA},
  year={2025},
  organization={IEEE}
}

@inproceedings{xiong2020push,
  title={Push and drag: An active obstacle separation method for fruit harvesting robots},
  author={Xiong, Ya and Ge, Yuanyue and From, P{\aa}l Johan},
  booktitle={ICRA},
  year={2020},
  organization={IEEE}
}

@article{jacob2024learning,
  title={Learning to simulate tree-branch dynamics for manipulation},
  author={Jacob, Jayadeep and Bandyopadhyay, Tirthankar and Williams, Jason and Borges, Paulo and Ramos, Fabio},
  journal={IEEE RAL},
  volume={9},
  number={2},
  year={2024},
  publisher={IEEE}
}

@inproceedings{rijal2025force,
  title={Force aware branch manipulation to assist agricultural tasks},
  author={Rijal, Madhav and Shrestha, Rashik and Smith, Trevor and Gu, Yu},
  booktitle={IROS},
  year={2025},
  organization={IEEE}
}

@inproceedings{jacob2024gentle,
  title={Gentle manipulation of tree branches: A contact-aware policy learning approach},
  author={Jacob, Jay and Cai, Shizhe and Borges, Paulo Vinicius Koerich and Bandyopadhyay, Tirthankar and Ramos, Fabio},
  booktitle={CoRL},
  year={2024}
}

@inproceedings{liu2024diff,
  title={Diff-control: A stateful diffusion-based policy for imitation learning},
  author={Liu, Xiao and Zhou, Yifan and Weigend, Fabian and Sonawani, Shubham and Ikemoto, Shuhei and Amor, Heni Ben},
  booktitle={IEEE/RSJ IROS},
  year={2024},
  organization={IEEE}
}

@inproceedings{ma2024hierarchical,
  title={Hierarchical diffusion policy for kinematics-aware multi-task robotic manipulation},
  author={Ma, Xiao and Patidar, Sumit and Haughton, Iain and James, Stephen},
  booktitle={2024 IEEE/CVF CVPR},
  year={2024},
  organization={IEEE}
}

\end{document}